\documentclass{article}

\usepackage{arxiv}

\usepackage[utf8]{inputenc} 
\usepackage[T1]{fontenc}    
\usepackage{hyperref}       
\usepackage{url}            
\usepackage{booktabs}       
\usepackage{amsfonts}       
\usepackage{amsmath} 
\usepackage{amssymb}
\usepackage{nicefrac}       
\usepackage{microtype}      
\usepackage{lipsum}
\usepackage{graphicx}
\graphicspath{ {images/} }
\usepackage{biblatex} 
\usepackage[dvipsnames]{xcolor}

\title{Larger than memory image processing}

\author{
 Jon Sporring\\
  Department of Computer Science\\
  University of Copenhagen\\
  Copenhagen, Denmark \\
  \texttt{sporring@di.ku.dk} \\
   \And
 David Stansby \\
  Department of Mechanical Engineering\\
  University College London\\
  London, United Kingdom \\
  \texttt{d.stansby@ucl.ac.uk} \\
}

\begin{document}
\maketitle
\begin{abstract}
This report addresses larger-than-memory image analysis for petascale datasets such as 1.4 PB electron-microscopy volumes \cite{shapson-coe.ea24} and 150 TB human-organ atlases \cite{walsh.ea21}. We argue that performance is fundamentally I/O-bound rather than compute-bound. We show that structuring analysis as streaming passes over data is crucial. For 3D volumes, two representations are popular: stacks of 2D slices (e.g., directories or multi-page TIFF) and 3D chunked layouts (e.g., Zarr/HDF5). While for a few algorithms, chunked layout on disk is crucial to keep disk I/O at a minimum, we show how the slice-based streaming architecture can be built on top of either image representation in a manner that minimizes disk I/O. This is in particular advantageous for algorithms relying on neighbouring values, since the slicing streaming architecture is 1D, which implies that there are only 2 possible sweeping orders, both of which are aligned with the order in which images are read from the disk. This is in contrast to 3D chunks, in which any sweep cannot be done without accessing each chunk at least 9 times. We formalize this with sweep-based execution (natural 2D/3D orders), windowed operations, and overlap-aware tiling to minimize redundant access.  Building on these principles, we introduce a domain-specific language (DSL) that encodes algorithms with intrinsic knowledge of their optimal streaming and memory use; the DSL performs compile-time and run-time pipeline analyses to automatically select window sizes, fuse stages, tee and zip streams, and schedule passes for limited-RAM machines, yielding near-linear I/O scans and predictable memory footprints.  The approach integrates with existing tooling for segmentation and morphology but reframes pre/post-processing as pipelines that privilege sequential read/write patterns, delivering substantial throughput gains for extremely large images without requiring full-volume residency in memory. 
\end{abstract}


\section{Introduction}
Recent advances in tomographic imaging at synchrotrons and in electron microscopy produce images that challenge even high-end workstations that have $\sim$1 TB RAM. For example, synchrotron volumes at DanMAX (\url{https://www.maxiv.lu.se/}) produce images with size $\sim$320 GB, Human Organ Atlas hub scans with size $\sim$2 TB, and the upcoming rotation stage for large volumes, which are estimated to generate images on the order of $\sim$150 TB \cite{esrf_bm18}. In electron microscopy, state-of-the-art volumes reach $\sim$1.4 PB \cite{shapson-coe.ea24}. These sizes dwarf local memory, making I/O-efficient, streaming-based processing essential. In this report, we will use Human Organ Atlas images as an example \cite{walsh.ea21}.

\subsection{Image from the Human Organ Atlas Hub}
\begin{figure}
    \centering
    \includegraphics[width=0.4\linewidth]{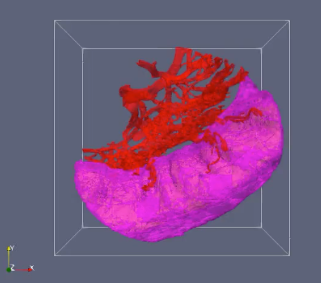}
    \caption{The segmentation of several structures in a zoom in of the human colon image at the HOAHub.}
    \label{fig:segmentation}
\end{figure}

The Human Organ Atlas Hub at the European Radiation Synchrotron Facility (ESRF) \cite{walsh.ea21} is built around HiP-CT, hierarchical phase-contrast tomography, which leverages the ESRF's Extremely Brilliant Source to image intact human organs non-destructively across scales, from whole-organ overviews ($\approx 25 \mu$m/voxel) down to organotypic units and even specialized cells in targeted volumes of interest (VOI) at $\approx 1 \mu$m / voxel. The approach couples careful organ stabilization/mounting with a multi-resolution, local-tomography workflow, enabling consistent image quality throughout large soft tissues and producing datasets that can reach hundreds of gigabytes for a single high-resolution VOI, underscoring the scale of data the hub routinely handles. An example of a segmentation of the Colon imaged at the HOAHub is given in Figure~\ref{fig:segmentation} 

Developed initially on the BM05 beamline, HiP-CT has benefitted further from ESRF’s BM18 beamline, designed for higher speed and sensitivity over volumes several times larger than human organs. Looking ahead, the hub plans to add a rotation stage that will enable true 3D scanning of a full human body at micron resolution; such acquisitions are expected to yield images on the order of ~150 TB, reinforcing that data management and I/O are important considerations for downstream analysis. 

\subsection{Image pipeline}

\begin{figure}
    \centering
    \includegraphics[width=0.8\linewidth]{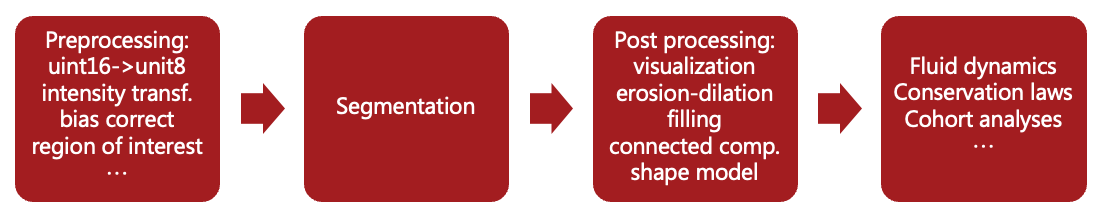}
    \caption{A common image processing pipeline}
    \label{fig:pipeline}
\end{figure}

"Image pipeline" is a loose term used to describe a process starting with an image, in this context 3-dimensional, followed by a number of steps until knowledge of some kind emerges. In many pipelines, segmentation is a central part, which is the task of classifying the individual voxels as either belonging or not to the object being segmented. An illustration of a typical pipeline is given in Figure~\ref{fig:pipeline}.
The steps are typically:
\begin{description}
    \item[Preprocessing] in a sense is a model of the imaging device, which translates the stored data to a form which is useful for the following measurement and analysis step. This often includes normalization of intensities, resizing images, and cropping regions of interest (ROI).
    \item[Segmentation] is a common pixel/voxel classification stage. It is not always performed, for example, analysis of directions using the structure tensor does not require segmentation, but for analyses at object level, segmentation is necessary, which is why, we have included it here.
    \item[Post-processing] in the case of, e.g., segmentation almost always require a post cleaning stage such as removing irrelevant segments and filling holes
    \item[Physical \& biological modelling] is here used as an example of the goal of the analysis. E.g., the clean segmentations are feed directly into a quantitative analyses, vessel graph extraction, lumen/mucosa metrics, or organ-scale transport models.
\end{description}

A significant cost of computing on very large images is reading and writing images from and from disk. Table~\ref{tab:ioSpeeds} illustrates practical constraints across memory, disk, and internet: even modest speeds translate to days–years to linearly scan petascale images, underscoring why streaming, single-pass I/O patterns are essential. 
\begin{table}
\centering
\caption{Indicative resource constraints for end-to-end linear scans of petascale images.}
\label{tab:ioSpeeds}
\begin{tabular}{lcc}
\hline
 (Capacity, speed, time)& \textbf{Macbook M1} & \textbf{Workstation} \\
\hline
\textbf{Memory}   & 16 GB, 5 GB/s, 3.2 days  & 1.1 TB, 2 GB/s, 8.1 days \\
\textbf{Disk}     & 1 TB, 44 MB/s, 1 years   & 11 TB, 76 MB/s, 213 days \\
\textbf{Internet} & $\infty$, 10 MB/s, 4.4 years    & $\infty$, 35 MB/s, 1.3 years \\
\hline
\end{tabular}
\end{table}
To keep the disk IO to a minimum, we advocate for a streaming architecture, where the image voxels are read and written as few times as possible. Some algorithms are well-suited for streaming architectures, for example, squaring all pixel values requires no knowledge of neighbouring pixel values and can thus be streamed at a pixel level, if needed, and can be done in a single sweep. Local operations, such as 3D filtering, require neighbouring 2D images to be in memory ($x$,$y$), but the calculation can be organised in a rolling window in $z$, thus requiring only a single sweep, under the assumption that the window can fit in memory. An overview of our classification is given in Table~\ref{tab:streaming-needs}.
\begin{table}
\centering
\caption{Streaming needs of common image-processing algorithms for larger-than-memory volumes. Window refers to the number of 2D slices that must be resident during a sweep of a slice stack.}
\label{tab:streaming-needs}
\begin{tabular}{p{3cm}p{4cm}p{1.8cm}p{1.5cm}p{4cm}}
\hline
\textbf{Algorithm class} & \textbf{Examples} & \textbf{$z$-window} & \textbf{Sweeps} & \textbf{Notes} \\
\hline
Single-pixel &
Intensity algebra, masking, threshold &
1 &
1 &
Trivially streaming. \\

Local neighbourhood &
Linear filtering, median filtering, mathematical morphology, non-maximum suppression &
$k$ &
1 &
Sliding window of size $k$ in stack direction. \\

Geometric transforms &
Axis permutation, re-slicing, affine transformation &
1 &
1-2 &
Transformations in the $x$-$y$ plane requires a single sweep, but efficient 3D transformation is most efficiently done in the chunked file structure, hence a slice to chunk transformation may be needed. \\

Global neighbourhood &
Connected components, Fourier transformation &
$1$--$k$ &
$>1$ &
Typically two-phase: E.g., Fourier transformation is calculated in $x$-$y$ first and then in $z$ in 2 or more sweeps. \\

Global reductions &
Histogram, pixel statistics, energy norms &
1 &
$\leq 1$&
E.g., histograms are calculated slice-wise and added between slices. Quantities such as the mean pixel value can be estimated by sampling slices. \\

Iterative solvers &
Partial Differential Equations, global graph cuts &
$1$--$k$ &
$\gg 1$ &
I/O-dominated on petascale and surrogates, separable/staged approximations, or reformulations is to be preferred. \\
\hline
\end{tabular}
\end{table}

In the following, we will present our Domain Specific Language (DSL), which allows the user to specify memory lean pipelines without considering the underlying streaming nature of the problem.

\subsection{Organising Images on Disk as Stacks versus Chunks}
\label{sec:diskOrganiztion}
Humans generate far more 2D images than 3D images, meaning technology for storing 2D images is extremely well studied and advanced. Storing 3D images as 2D slices allows us to take advantage of existing 2D data formats and processing tools. In contrast, newer formats for 3D images, such as Zarr \url{https://zarr.dev/]} or HDF5 \url{https://www.hdfgroup.org/solutions/hdf5/} formats, are still behind in popularity and tool support.

\begin{figure}
    \centering
    \includegraphics[width=0.3\linewidth]{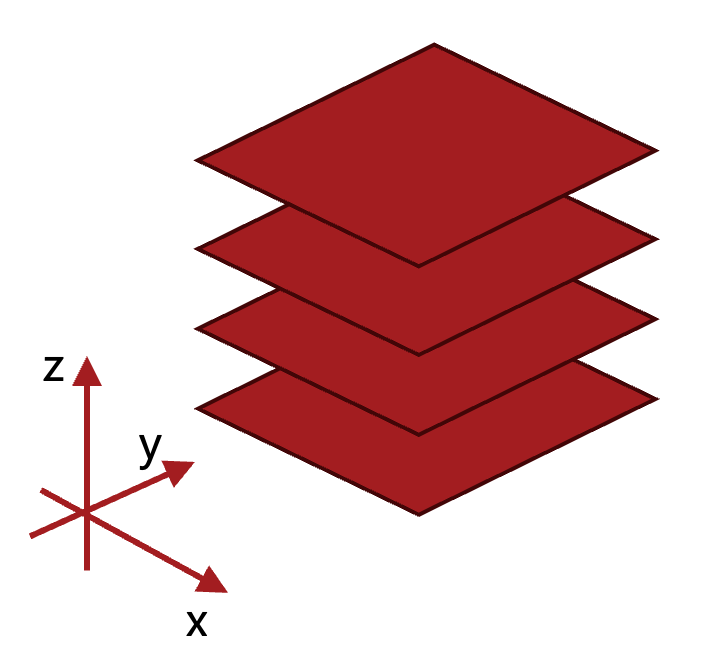}
    \includegraphics[width=0.3\linewidth]{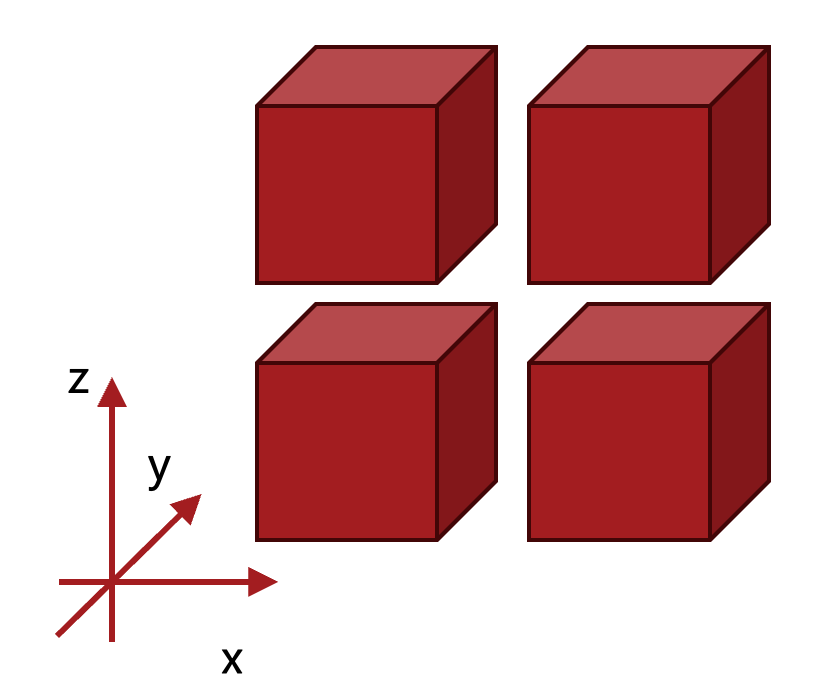}
    \caption{An illustration of representing a 3-dimensional image as a stack of 2-dimensional slices (LEFT) and as a set of non-overlapping 3-dimensional sub-images also called chuncks (RIGHT).}
    \label{fig:slicesVsChuncks}
\end{figure}

Stacks of images represent a 3D volume as an ordered series of 2D slices (e.g., one file per z-plane, or a multipage TIFF) as illustrated in Figure~\ref{fig:slicesVsChuncks}(LEFT).
This layout facilitates slice-wise streaming, keeps only a small window (a few neighbouring slices) in memory, and writes results out as you go. By contrast, Zarr \cite{zarrdev} stores n-dimensional arrays in chunked blocks inside a container as illustrated in Figure~\ref{fig:slicesVsChuncks}(RIGHT). Chunking is great for random access, out-of-core indexing, and cloud distribution (object storage, parallel reads, consolidated metadata). HDF5 \cite{hdf5} is a container that can hold both slice stacks and chunk collections. 

\begin{figure}
    \centering
    \includegraphics[width=0.8\linewidth]{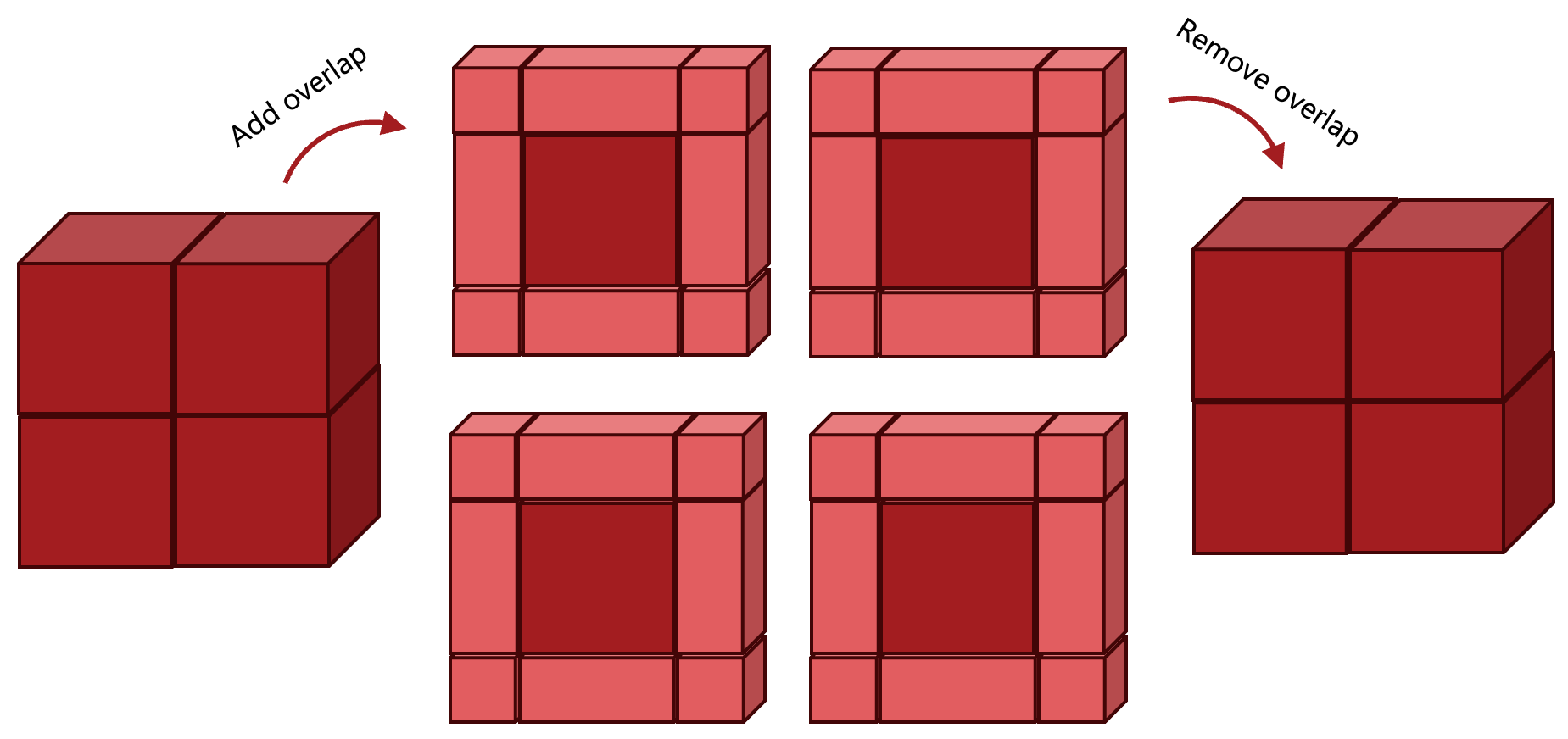}
    \caption{An illustration of the halo process: Each chunk to be processed is extended with parts of its neighbours, the computation is performed, and the center is extracted. The result is recombined into a resulting image.}
    \label{fig:halo}
\end{figure}

For processing very large images, the storage structure on disk and the sweeping order in memory need to be aligned. E.g., 3D filtering with a kernel of size $(m,n,o)$ needs at minimum neighbouring chunks of $m\times n\times o$ in memory in order to calculate the output of a single voxel without further I/O. In order to produce output of the size $m\times n\times o$, a chunk of size $(2m-1)\times (2n-1)\times (2o-1)$ of the original image needs to be accessed. These extra values are called the halo. An illustration of a halo process is shown in Figure~\ref{fig:halo}.
In 3D, a $(m,n,o)$ chunk will have 26 neighbouring chunks, which need to be read. If the chunks are treated in random order, this requires rereading of each chunk 26 times and represents the worst-case scenario for disk I/O access. Streaming chunks is, however, not an easy solution, since in 3D there are many curves in which the collection of blocks can be traversed. Most curves will have the effect that the 9 chunks in the direction of the curves will need to be read in order to move from the processing of one $3\times3\times3$ set of neighbouring chunks to the next, and this implies that every chunk is read 9 times for every sweep. In contrast, the 2D slice organization only has 2 curves (downward and upward in the $z$-direction), and hence, local or reduce algorithms can be performed in a single sweep, reading each image only once. These two approaches can be combined, that is, if the chunk size is limited such that two layers of all $x$--$y$ chunks can be stored in memory, then slices can dynamically be extracted from the chunks, and the chunks can be processed in a single sweep, similarly to the slice-wise organization. 

\subsection{Review of existing tools for parallel programming and streaming}

At petascale, the limiting factor is I/O and memory locality, so we distinguish frameworks that primarily parallelize chunked computations from those that stream data through bounded-memory pipelines. Some relevant frameworks are given below and summarized in Table~\ref{tab:tools-review}.
\begin{table}
\centering
\caption{Existing tools relevant to large-image analysis: primary emphasis and typical use.}
\label{tab:tools-review}
\begin{tabular}{p{0.2\linewidth}p{0.2\linewidth}p{0.5\linewidth}}
\hline
\textbf{Tool / Project} & \textbf{Primary focus} & \textbf{Notes} \\
\hline
ITK (streaming filters) & Streaming & Region-of-interest requests; bounded-memory filters.  \\
Dask (with xarray) & Parallelism & Chunked arrays, distributed scheduling, blockwise. \\
Cubed & Streaming \& parallel & Bounded-memory array processing; NumPy/xarray API; drop-in for Dask Array.  \\
Zarr & Storage (chunked) & Container enabling random access/cloud-native I/O; pairs with Dask/Cubed.  \\
HDF5 & Storage (chunked) & Single-file (or family) container with chunking/compression; similar trade-offs.  \\
\hline
\end{tabular}
\end{table}
ITK's streaming filters are a long-standing example of the latter in image processing, emphasizing slice/block-wise execution with explicit control over requested regions \cite{itk-book1}. In the array-computing ecosystem, Dask \cite{xarray_dask_guide_2024} focuses on parallel execution over chunked nD arrays and distributed scheduling, whereas Cubed \cite{cubed_docs} aims for bounded-memory, out-of-core array processing with compile-time checks for memory use and a drop-in API—well-suited to streaming-style pipelines. Finally, Zarr \cite{zarrdev} and HDF5 \cite{hdf5} are storage/container formats that provide chunked layouts enabling both approaches, but they are not execution engines by themselves; in our context they primarily influence access patterns and I/O costs. From a user point of view, optimal use of these frameworks is not easy, as, e.g., the ITK Software Guide states \cite[Chapter 8.3.1]{itk-book1}:
\begin{quote}
    "The secret to creating a streaming filter is to understand … how to override its default behaviour by using the appropriate virtual functions"
\end{quote}

\section{Memory model for larger-than-memory 3D operators}
Given a function
$$\beta=f(\alpha),$$
where $\alpha$ and $\beta$ typically will be array-types. We favour the functional programming perspective, where $\alpha$ and $\beta$ are constants and there are no side-effects. Thus, the storage requirement for evaluating it is
\begin{equation}
    M\{f\} = |\alpha|+|\beta|+|\gamma|,
    \label{eq:memoryModel}
\end{equation}
where where $|x|$ is the memory size of the data structure $x$, and $|\gamma|$ is the size of internal variables. If there are only $m$ bytes available for the computation, then
\begin{equation}
    M\{f\} < m
    \label{eq:memLimit}
\end{equation}
ie. the memory requirement must be limited. From an imperative perspective, some functions can be programmed as update functions, e.g., $\alpha\rightarrow \alpha + 1$, and the memory requirement can be reduced to $|\alpha|+|\gamma|$, however, all functions which depend on neighbouring values such as convolution cannot be computed as an update function, and thus the functional perspective is a simpler model for reasoning about the memory requirements. Further, the functional perspective allows for sharing of the input $\alpha$ between two independent functions, and therefore will only need a single copy of $\alpha$ in memory regardless of the number of functions using it as input.

\subsection{Sweeping and halos}
As discussed in Section~\ref{sec:diskOrganiztion}, the organisation of the image on disk and its interface to a slice-based streaming architecture can be aligned. Hence and unless otherwise stated, we will in the following consider a slice-wise streaming architecture in the following.

Many 3D operators (e.g., $k^3$ convolutions, morphology, partial differential equation (PDE) stencils) require halo voxels around each processing region. When volumes are larger than memory, the dominant cost becomes moving bytes rather than arithmetic, and the data layout/scheduling determines how often the same bytes are reread from storage. Representing the volume as a stack of 2D slices enables a 1D sweep (e.g., along the $z$-direction) that keeps only a narrow sliding window of slices in memory. For a convolution kernel with edge sizes $k_x$, $k_y$, and $k_z$, a single pass requires a sliding window of $k_z$ consecutive images to produce the center, valid slice. For simplicity, we will assume that $k=k_x=k_y=k_z$ such that the kernel consists of $k^3$ voxels. The sweep advances by one slice at a time, expelling the oldest slice and reading exactly one new slice. Consequently, a single sweep reads the data exactly once. However, slice-wise sweeps limits the maximum width of a 3d image which can be processed: For example, consider a cubic image with size $n^3$ voxels of size $b$ bytes, then to support a single-pass, slice-wise sweep with a $k$-slice window, which produces 1 slice, memory must satisfy
\begin{equation}
    M\{f\} = (k+1)n^2b + \mathcal{O}(1),
    \label{eq:memoryConvSingle}
\end{equation}
where $\mathcal{O}(1)$ is a small value accounting for the internal storage needed for performing the computation independent on the image and kernel sizes. Taking \eqref{eq:memLimit} into account, the maximum image width will then approximately be,
$$
n \approx \sqrt{\frac{m}{k+1}}.
$$
For example, if voxels are bytes, $m=1~\mathrm{TB}=2^{40}$, and $k=10$, then
\[
n \approx \sqrt{\frac{2^{40}}{11}}\approx
316\,158,
\]
in which case the maximum size of the volume, $n^3$, is  approximately 28 PB. As a side note, there are no limit to the size of the volume in terms of the number of slices as long as $n \lesssim 316\,158$

\subsection{Higher-order functionals on stacks}
In general, for a 3D input volume stored as a stack of 2D images, algorithms can be characterized by the by how many slices is needed for a step in a sweep and how many sweeps are needed to perform the computation. For example to calculate the histogram of an image, all slices need read once, but an estimate of the histogram may be computed from reading only a subset of slices. Solving a partial differential equation using a finite difference scheme requires a small sweeping window but a large number of sweeps. Finally, functions such as computing the connected components in a single sweep requires the complete image to read into memory, and this can be considered a windowed operation, where the window size is equal to number of images in the stack. For each sweep, we focus on a small set of higher-order functionals for stacks consisting of elements, and the elements can be anything but in our case typically slices or stacks of slices. These are:
\begin{description}
    \item[\texttt{windowed}:] Make a stack of stacks, which are sliding window of given size, stride, padding, e.g., for window size $w$, stride $s$ and padding $p=0$ this can be illustrated as 
    $$(w,s,p,[e_0,e_1,e_2,\ldots]) \rightarrow [[e_0,e_1,\ldots,e_{w-1}], [e_s,e_{s+1},\ldots,e_{s+w-1}], \dots].$$
    \item[\texttt{flatten}:] Flatten the stacks of stacks into a stack by list concatenation, e.g., 
    $$[[e_0,e_1], [e_2,e_3], \dots]\rightarrow [e_0,e_1,e_2,\ldots].$$
    \item[\texttt{map:}] Apply a function to each element and return the list of the resulting values, e.g., 
    $$(f,[e_0,e_1,e_2, \dots])\rightarrow [f(e_0),f(e_1),f(e_2),\ldots].$$
    \item[\texttt{fold}:] Update an accumulator as 
    $$(a_0, [e_0,e_1,e_2, \dots])\rightarrow a_n,$$
    where $a_{i+1} = f(a_i, e_i)$ is a function, which updates the accumulator.
    \item[\texttt{zip}:] Combine two streams to make a stream of pairs,
    $$([e_0^0,e_1^0,e_2^0, \dots],[e_0^1,e_1^1,e_2^1, \dots]) \rightarrow [[e_0^0,e_0^1],[e_1^0,e_1^1],[e_2^0,e_2^1],\ldots].$$
    \item[\texttt{initialize}:] Generate a stack of a given depth,
    $$(d,f) \rightarrow [f(0),f(1),\ldots,f(d-1)].$$
\end{description}
Common for all but \texttt{initialize}, is that they transform streams, and although many languages implement garbage collection, it is our experience, that garbage collectors are not eager enough for keeping the memory consumption at an absolute minimum. Hence, we reduce the memory pressure by reference count semantics, that is, each time an element is part of an output, then its reference count is increased by 1, when an element is consumed then its reference count is reduced by 1, and when the reference count for an element reaches zero, then its memory is released. 

The memory pressure of the functionals independently depends on the computation. Both \texttt{map} and \texttt{fold} are conveniently composed with \texttt{windowed}. For example, function $f$ whose computation requires a local 3D neighbourhood, such as convolution can be transformed into streaming function using windowing as follows,
\begin{quote}
    \texttt{windowedStream$_f$ = flatten $\circ$ map$_f$ $\circ$ windowed$_{w,s,p}$}.
\end{quote}
For any windowed function on a flat stream, in the notation of \eqref{eq:memoryModel}, $|\alpha|=wn^2b$, and the size of $\beta$ and $\gamma$ depends on the specific algorithm. For example, if \texttt{f $=$ convolve$_K$}, where $K$ is the kernel with $k_z$ as the size in the stacking direction. For convolutions, we must use $w\geq k_z,$ to ensure that the resulting center slices are independent on any boundary conditions, then there are $w-k_z+1$ such center slices, and $|\beta| = (w-k_z+1)n^2b$. Thus, 
\begin{equation}
    M\{\texttt{convolve}_K\} = (2w-k_z+1)n^2b + \mathcal{O}(1),
\end{equation}
assuming $|\gamma|$ is negligible. Note that in this case, to avoid repeated slices for convolution, we must have $s=w-k_z+1$. If $f(I(x,y,z))$ is a pointwise computation, such as $f(I(x,y,z))=I(x,y,z)^2$, then this can also be computed as a \texttt{windowedStream$_f$} function with $w>0$, In case $w=1$, then this solution has a small but insignificant computational overhead. In order to apply $f$ to all voxels of a flat stream and assuming that the input and output types are identical, $|\alpha|=|\beta| = wn^2b$, and 
\begin{equation}
    M\{\texttt{square}\} = 2wn^2b + \mathcal{O}(1).
\end{equation}

An example of a function whose computation folds the stack is the histograms of an image, since the histogram of a stack of two images is the sum of the histogram of each of them individually. For such computations can be transformed into a streaming function as 
\begin{quote}
    \texttt{foldStream$_f$ = fold$_{a_0,(+)}$ $\circ$ map$_f$ $\circ$ windowed$_{w,s,p}$}.
\end{quote}
where $(+)$ is the function, which updates the accumulator. When \texttt{f $=$ histogram}, then $(+)(a, b) = a + b$ is the addition function, $w>0$. Similarly to pointwise functions above, if $w=1$ then this solution has a small but insignificant computational overhead. The memory requirement for folding a flat stream is $|\alpha| = wn^2b$, and for the histogram function, an upper bound on the size of the histogram is $256^b$, in which case we need to store the old and the new accumulator together with the histogram of a window. Here, however, we may reduce the memory footpint a little by iteratively updating the output histogram, such that only the old and the new need be saved, hence
\begin{equation}
    M\{\texttt{histogram}\} = wn^2b+2\cdot 256^b + \mathcal{O}(1).
\end{equation}
Since \texttt{fold} consumes a stream, it acts similarly to a sink.

The \texttt{zip} is typically used in combination with \texttt{map}. For example, to add two images of flat streams, one can \texttt{add($I$,$J$) $=$ map$_{(+)}$ zip($I$,$J$)}, where $(+)$ is the function which adds two slices in a stack. The zipper needs a single slice of each stream once, and since there is no computation done, the output does not require further memory and 
\begin{equation}
    M\{\texttt{zip}\} = 2n^2b + \mathcal{O}(1).
\end{equation}

Finally, \texttt{initialize} is in a category of its own, since it does not transform streams but generates data and as such acts as a source. For example, \texttt{initialize} can be used to generate a constant image, e.g., $I(x,y,z)=0$. For this, $|\alpha|=0$ and $|\beta| = n^2b$, and
\begin{equation}
    M\{\texttt{zero}\} = n^2b + \mathcal{O}(1).
\end{equation}

\subsection{Sequential and parallel composition}
When composing a larger-than-memory pipeline from streaming operators with sliding windows, the dominant constraint is RAM. When two windowed stages are concatenated either sequentially or in parallel branches, their instantaneous memory footprints add.

A sequential image pipeline consists of the composition of functions, $f(g(I))$, where $I$ is the image and $f$ and $g$ are the two streaming functions. The memory usage is additive: When $g$ transforms its stream, it uses $M\{g\}=|\alpha_g|+|\beta_g|+\mathcal{O}(1)$ memory. As the output of $g$ is streamed to $f$, then $f$ uses $M\{f\}=|\alpha_f|+|\beta_f|+\mathcal{O}(1)$ memory, 
\begin{equation}
    M\{f\circ g\} \leq M\{f\}+M\{g\}.
\end{equation}
For some pipelines, the output of $g$ is used as it is produced, e.g., two consecutive convolutions, where the window sizes are equal to the kernel depths, and in such cases $M\{f\circ g\} = M\{f\}+M\{g\}-|\beta_g|$. However, in other case, e.g, when $g$ outputs a large number of slices per iteration and $f$ is a window over a few slices, then $g$'s output cannot be released before $f$ has processed them, so there is no memory to be save. An example of where memory can be reduced, is the composition of two convolution operators with kernels $K$ and $L$ of size $k^3$ and $l^3$ with window sizes equally $k$ and $l$. As the first streaming process is producing images requiring memory for $k+1$ slices, the second needs to sweep the output of the first requiring memory for $l+1$ slices, hence the memory usage of the composition is $M\{\texttt{convolve}_K(\texttt{convolve}_L(I))\}\approx (k+l+2)n^2b$. Note that particularly for composition of convolution operators, this is wasteful, since convolution is associative, so instead of consecutive convolutions, we may convolve the image with a single convolution $\texttt{convolve}_{K*L}$. The resulting kernel will have side-lengths $k+l-1$, and since we still need to store the slice-wise result, we would save memory for the allocation of two slices. For general sweeping operations this is not the case.

For parallel pipelines on the image: $[f(I),g(I)]$, when $f$ and $g$ are computed independently on each other, the combined memory requirement is again additive.
\begin{equation}
    M\{[f(I),g(I)]\} \leq M\{f\}+M\{g\}.
\end{equation}
For example, for the convolution with kernel depths $k+1$ and $l+1$ slices in memory and the combined will be $k+l+2$. However, in many cases, we can synchronize the sweep. For example, if $l<k$ then the sweeping implementation $g$ can be changed to be the convolution of a window size $k$ with stride $s=k-l+1$. Then they can share the input sweeping window, and the required memory becomes $k+1+k-l+1=2k-l+2$. For example if $k=l$ then only $k+2$ slices need to be stored in  memory at each sweeping step. 

\subsection{Optimising pipelines}
We have identified two simple optimisation levers for handling the memory pressure in sequential and parallel pipelines: 
\begin{enumerate}
    \item introduce a mid-write to complete a build-up without exceeding memory.
    \item jointly optimise window sizes to minimise total time under a memory cap.
\end{enumerate}
\paragraph{Mid-write splits:} Assuming that there is disk space available to write intermediate results of the pipeline, a sequential pipeline consisting of a number of functions $f_{q-1}(f_{q-2}(\dots(f_0)\dots))$ can be split into several pipelines with intermediate results written and read from disk. Consider the memory pressure of the non-modified pipeline:
$$
    M\{f_{q-1}(f_{q-2}(\dots(f_0)\dots))\} \leq \sum_{i=0}^{q-1} M\{f_i\}
$$
The first function $f_0$ will a source such as reading slices from disk, and assuming that the total memory budget $m>M\{f_0\}+M\{f_1\}$, then there we may find the largest value $j$ such that $m>M\{\texttt{write}\}+\sum_{i=0}^{j} M\{f_i\}$ at which point we can insert a mid-write and read point: $\texttt{write}(f_{j}(f_{j-1}(\dots(f_0)\dots)))$ followed by $f_{q-1}(f_{q-2}(\dots(f_{j+1}(\texttt{read}))\dots))$.  This splits the pipeline into two memory wise, 
$$
    M\{f_{q-1}(f_{q-2}(\dots(f_0)\dots))\} \leq \max \Big(M\{\texttt{read}\}+\sum_{i=j+1}^{q-1} M\{f_i\}, \quad M\{\texttt{write}\}+\sum_{i=0}^{j} M\{f_i\}\Big)
$$
By recursion, we can complete the pipeline within the memory budget under the assumption that at each split, a largest $j>0$ exists, $m>M\{\texttt{read}\}+M\{f_{j+1}\}$, and temporary disk storage is available for each split.

\paragraph{Window-size optimization:}
For algorithm allowing for sliding window processing, the size of the sliding window may be any value between a given algorithm's minimum size to the total number of slices in the stack. For example, the convolution with a kernel with depth $k$, then the window size is any value $w$: $k\leq w \leq n$, where $n$ is the number of slices in the 3-dimensional volume. For optimization purposes, the running time for each algorithm becomes an

Assuming that the computation of a sliding window function $f_i$ is faster, the bigger a window it operates on, we have the optimization problem
\begin{align*}
&\max_{W} T_W \quad \text{s.t.} \quad m > T_W\\
&T_W=\sum_{i=0}^{q-1} M\{f_i^{w_i}\}   
\end{align*}
where  $W=[w_0,w_1,\dots]$ are the (possible) window-size parameters of the corresponding function $f_i.$

\section{stackProcessing (FSharp.StackProcessing).}
The above an more have been implemented in a domain specific language, we call \emph{stackProcessing} (\url{https://github.com/sporring/stackProcessing}), which is an F\# library for building memory-efficient, streaming pipelines over large 3D image stacks. It wraps a functional dataflow (\texttt{SlimPipeline}) where you declare a \texttt{source} with a memory budget, compose \texttt{Plans} into a \texttt{Pipeline} using operators to compose, fanout and fan in pipeline branches, and finally optimize and execute with a \texttt{sink}. An overview is as follows:
\begin{description}
    \item[Core abstractions]~\\[-1em]
\begin{itemize}
  \item \texttt{Source} / \texttt{Sink}: streaming endpoints that read/write slice stacks.
  \item \texttt{Plan}, \texttt{State} and \texttt{Pipe}: a plan describes a stage; a pipeline composes stages under a memory budget.
  \item \texttt{Budget}: explicit bound on in-flight memory (slices, windows, buffers).
\end{itemize}
    \item[Basic I/O]~\\[-1em]
\begin{itemize}
  \item \texttt{read}: stream slices from a directory or multipage file.
  \item \texttt{readInChunks}: stream slices from a directory or multipage file.
  \item \texttt{write}: write results as a slice stack (or other sink).
  \item \texttt{writeInChunks}: write results as a slice stack (or other sink).
\end{itemize}
    \item[Sources, sinks and basic streaming operators]~\\[-1em]
\begin{itemize}
  \item \texttt{source}: Initial point of a pipeline with a memory budget.
  \item \texttt{sink}: End point of a pipeline with memory optimization and budget checks.
  \item \texttt{window}: expose a $(2k{-}1)$--slice sliding window for 3D context.
  \item \texttt{flatten}: collapse windowed output back to a simple slice stream.
  \item \texttt{map}: apply a per 2D image or window-wise function.
  \item \texttt{fold}: reduce a stream by interatively updating an accumulator based on the streamed 2D images or windows.
\end{itemize}
    \item[Local neighbourhood (sliding--window) operators]~\\[-1em]
\begin{itemize}
    \item\texttt{medianFilter}, \texttt{erode},  \texttt{dilate}, \texttt{discreteGaussian}:
        local filters that consume a window and emit a slice stream.
  \item \texttt{threshold t}: common pointwise ops.
\end{itemize}
    \item[Geometric / structural transforms]~\\[-1em]
\begin{itemize}
  \item \texttt{permuteAxes xyz$\rightarrow$zyx}, \texttt{reslice axis}: reordering/reslicing as pure streaming.
  \item \texttt{crop}, \texttt{pad}: streaming edits of image extents.
\end{itemize}
    \item[Composition and control]~\\[-1em]
\begin{itemize}
  \item \verb|>=>}|(append): chain stages left--to--right as a single stream.
  \item \verb|-->| (compose): function--style composition of plans.
  \item \verb|>=>>| (tee/branch): split a stream into parallel branches.
  \item \verb|>>=>| (join/zip): rejoin or combine branch outputs slice--aligned.
  \item \texttt{zip f}: combine two synchronized streams with function \texttt{f}.
\end{itemize}
\end{description}
A typical programming pattern is as follows:
\begin{verbatim}
source memBudget
  >=> read "inputDir"
  >=> discreteGaussian 1.5
  >=> write "outDir"
  |> sink
\end{verbatim}
This example reads a slice stack, applies a bounded 3D filter via a sliding window and writes the result, all as a single pass whose memory footprint is controlled by the explicit window size and the configured budget.

\section{Conclusion}
Slice-wise streaming architectures provide the most practical path to petascale image analysis: they read each slice exactly once (or a few times), bound memory to a small sliding window, and avoid the halo rereads inherent to 3D chunk traversal. This strategy works naturally with classic stack layouts (one file per slice or multi-page TIFF) and also adapts well to Zarr-style stores by scanning chunked planes sequentially and minimizing random access. A few operations, notably large geometric resampling and global Fourier transforms, benefit from (or temporarily require) chunk-based layouts to achieve acceptable locality and throughput; these can be handled as bounded, format-aware checkpoints within an otherwise streaming pipeline. Despite the need to keep an entire $(k\times n \times n$ slice window resident, the approach scales: with $\sim$1~TiB of RAM, windowed filters can process two-digit petabyte volumes in single/few sweeps, delivering predictable performance on commodity workstations while preserving I/O efficiency at extreme data sizes.

\printbibliography 

@misc{esrf_bm18,
  author       = {{European Synchrotron Radiation Facility (ESRF)}},
  title        = {BM18 Beamline Overview},
  howpublished = {\url{https://www.esrf.fr/home/UsersAndScience/Experiments/StructMaterials/BM18/over.html}},
  year         = {2025},
  note         = {Accessed: 2025-10-20}
}

@article{shapson-coe.ea24,
author = {Alexander Shapson-Coe  and Michał Januszewski  and Daniel R. Berger  and Art Pope  and Yuelong Wu  and Tim Blakely  and Richard L. Schalek  and Peter H. Li  and Shuohong Wang  and Jeremy Maitin-Shepard  and Neha Karlupia  and Sven Dorkenwald  and Evelina Sjostedt  and Laramie Leavitt  and Dongil Lee  and Jakob Troidl  and Forrest Collman  and Luke Bailey  and Angerica Fitzmaurice  and Rohin Kar  and Benjamin Field  and Hank Wu  and Julian Wagner-Carena  and David Aley  and Joanna Lau  and Zudi Lin  and Donglai Wei  and Hanspeter Pfister  and Adi Peleg  and Viren Jain  and Jeff W. Lichtman },
title = {A petavoxel fragment of human cerebral cortex reconstructed at nanoscale resolution},
journal = {Science},
volume = {384},
number = {6696},
pages = {eadk4858},
year = {2024},
doi = {10.1126/science.adk4858},
URL = {https://www.science.org/doi/abs/10.1126/science.adk4858}
}

@article{walsh.ea21,
	author = {Walsh, C. L. and Tafforeau, P. and Wagner, W. L. and Jafree, D. J. and Bellier, A. and Werlein, C. and K{\"u}hnel, M. P. and Boller, E. and Walker-Samuel, S. and Robertus, J. L. and Long, D. A. and Jacob, J. and Marussi, S. and Brown, E. and Holroyd, N. and Jonigk, D. D. and Ackermann, M. and Lee, P. D.},
	month = jan,
	doi = {10.1038/s41592-021-01317-x},
	journal = {Nature Methods},
	number = {12},
	pages = {1532--1541},
	title = {Imaging intact human organs with local resolution of cellular structures using hierarchical phase-contrast tomography},
	url = {https://doi.org/10.1038/s41592-021-01317-x},
	volume = {18},
	year = {2021},
}

@misc{zarrdev,
  author       = {{Zarr Developers}},
  title        = {Zarr: Open Standard for Chunked, Compressed N-Dimensional Arrays},
  howpublished = {\url{https://zarr.dev/}},
  year         = {2025},
  note         = {Accessed: 2025-10-20}
}

@misc{hdf5,
  author       = {{The HDF Group}},
  title        = {Hierarchical Data Format, Version 5 (HDF5)},
  howpublished = {\url{https://www.hdfgroup.org/}},
  year         = {2025},
  note         = {Accessed: 2025-10-20}
}

@misc{xarray_dask_guide_2024,
  author       = {{xarray developers}},
  title        = {User Guide/Parallel computing with Dask},
  howpublished = {\url{https://docs.xarray.dev/en/v2024.10.0/user-guide/dask.html}},
  year         = {2024},
  note         = {xarray User Guide v2024.10.0. Accessed: 2025-10-20}
}

@misc{cubed_docs,
  author       = {{Cubed Developers}},
  title        = {Scalable array processing with bounded memory --- Cubed},
  howpublished = {\url{https://cubed-dev.github.io/cubed/}},
  year         = {2025},
  note         = {Accessed: 2025-10-20}
}

@incollection{itk-book1,
  author    = {Hans J. Johnson and Matthew M. McCormick and Luis Ib\'{n}\tilde{e}ez, and {the Insight Software Consortium}},
  title     = {The ITK Software Guide, Book 1: Introduction and Development Guidelines},
  booktitle = {The ITK Software Guide},
  edition = {4},
  publisher = {Insight Software Consortium},
}

\end{document}